\documentclass{article} 

\usepackage{xcolor}
\usepackage{ulem}
\usepackage{booktabs}
\usepackage{tcolorbox}
\usepackage{makecell}

\definecolor{accGreen}{RGB}{21, 128, 61}
\definecolor{rejRed}{RGB}{185, 28, 28}
\definecolor{corBlue}{RGB}{29, 78, 216}
\definecolor{tagBg}{RGB}{241, 245, 249}
\definecolor{boxBorder}{RGB}{203, 213, 225}

\newcommand{\acc}[1]{\textcolor{accGreen}{\texttt{#1}}}
\newcommand{\rej}[1]{\textcolor{rejRed}{\sout{\texttt{#1}}}}
\newcommand{\cor}[1]{\textcolor{corBlue}{\textbf{\texttt{#1}}}}
\newcommand{\arrow}{\;$\rightarrow$\;}

\usepackage{xcolor}
\usepackage{booktabs}
\usepackage{tabularx}
\usepackage{tcolorbox}
\usepackage{soul} 

\definecolor{acceptGreen}{RGB}{20, 125, 45}
\definecolor{rejectRed}{RGB}{185, 28, 28}
\definecolor{correctBlue}{RGB}{29, 78, 216}
\definecolor{bonusPurple}{RGB}{109, 40, 217}
\definecolor{roundBorder}{RGB}{203, 213, 225}
\definecolor{roundBg}{RGB}{248, 250, 252}

\usepackage{iclr2027_conference,times}
\iclrfinalcopy

\usepackage[bottom=1in, footskip=0.5in, left=1.0in,right=1.0in]{geometry}

\usepackage{amsmath,amsfonts,bm}

\def\eqref#1{equation~\ref{#1}}

\def\1{\bm{1}}

\DeclareMathAlphabet{\mathsfit}{\encodingdefault}{\sfdefault}{m}{sl}
\SetMathAlphabet{\mathsfit}{bold}{\encodingdefault}{\sfdefault}{bx}{n}

\usepackage{hyperref}
\usepackage{url}

\usepackage{amsmath,amssymb}
\usepackage{algorithm}
\usepackage{algpseudocode}
\usepackage{booktabs}
\usepackage{cite}
\usepackage{booktabs}
\usepackage{multirow}
\usepackage{graphicx}
\usepackage{makecell}
\usepackage{wrapfig}
\usepackage{enumitem}
\newcommand{\Lest}{\bar L_D}
\newcommand{\Hest}{\bar H_D}
\newcommand{\Iest}{\bar I_D}
\newcommand{\dest}{\bar d_D}

\title{\texttt{T2D}: Target-to-Draft Model Pruning \& Adaptation for Efficient Speculative Decoding}

\usepackage{authblk}

\author[1]{%
    Shahriar Kabir Nahin
}
\author[2]{%
     Xiao Ye%
}

\author[3]{%
     Muhao Chen%
}

\author[2]{%
     Ben Zhou%
}
\author[1]{%
    Anshuman Chhabra%
}
\affil[1]{Bellini College of AI, Cybersecurity, and Computing, University of South Florida}
\affil[2]{School of Computing and Augmented Intelligence, Arizona State University}
\affil[3]{Department of Computer Science, University of California, Davis}

\begin{document}

\maketitle

\begingroup
\renewcommand\thefootnote{}
\footnotetext{%
  \makebox[0.93\textwidth][c]{%
    \textbf{Emails:}~\textsuperscript{1}\texttt{\{shahriarkabir,anshumanc\}@usf.edu}
    \textsuperscript{2}\texttt{\{xiaoye2,xzhou202\}@asu.edu}
    \textsuperscript{3}\texttt{muhchen@ucdavis.edu}%
  }%
}
\endgroup

\begin{abstract}
\looseness-1 Speculative decoding significantly accelerates autoregressive text generation in Large Language Models (LLMs), but its efficacy depends largely on two key factors: the distributional alignment between the \textit{draft} and the \textit{target} model, and the \textit{draft} model's latency during inference. However, existing approaches typically utilize independently pretrained small draft models or seek to prune the target model to a manually chosen architecture, thereby failing to explicitly balance the trade-off between draft model alignment and token generation efficiency. To address this gap, we formulate draft model design as an \textit{optimization} problem and propose \texttt{T2D}, a framework that constructs draft models directly from the target model using \textit{latency-aware structured pruning} and \textit{knowledge distillation}. In this manner, \texttt{T2D} obtains draft models that are jointly optimized for both measured inference latency and cross-entropy loss on the target model's responses, resulting in superlative speculative decoding performance. Through extensive experiments on multi-domain speculative decoding benchmarks and across several target model families of differing scale, we demonstrate how \texttt{T2D} prunes up to 92\% of the target model to obtain highly performant draft models compared to competitive baselines. Moreover, draft models obtained via \texttt{T2D} can also be utilized by other speculative decoding token-alignment and knowledge distillation methods in a plug-and-play fashion, leading to further gains. Our findings underscore the importance of jointly navigating the draft model architecture-alignment frontier to unlock the full potential of speculative decoding for LLMs.

\end{abstract}

\section{Introduction}
\label{sec:introduction}

\looseness-1Large Language Models (LLMs) have become the foundation of many modern AI applications, including chat assistants, code generation tools, and document summarization systems~\citep{zhao2026survey, naveed2025comprehensive,askari-etal-2025-assessing,belay2026afrilangtutor, bonagiri2025towards}. As modern LLMs consist of billions of parameters, \textit{inference efficiency} has become an important consideration alongside ensuring high-quality response generation~\citep{zhou2024survey}. At the same time, improving response quality often requires larger models, making it challenging to maintain \textit{low latency} and reasonable \textit{serving costs} due to the ever increasing parameter scale \citep{lopez2026can}. This challenge has motivated a growing body of research on reducing the inference latency of autoregressive text generation in LLMs.

\textit{Speculative decoding}~\citep{leviathan2023fast} seeks to address this challenge without compromising output quality by incorporating a smaller model into the text generation process. Instead of generating every token with a large and high-performing model, a low latency model, known as the \textit{draft} model, first generates a short sequence of 4-5 tokens \citep{leviathan2023fast, chen2023accelerating}. The larger original model, also known as the \textit{target} or \textit{verifier} model, then verifies the drafted sequence in a single forward pass and accepts only the subsequently matching tokens. Since most tokens are generated by the low latency \textit{draft} model, and the larger model is used only for verification, speculative decoding can significantly reduce the latency of autoregressive decoding while preserving the output quality guaranteed by the \textit{verifier}. 

\begin{figure}[t]
    \centering
    \includegraphics[width=1\linewidth]{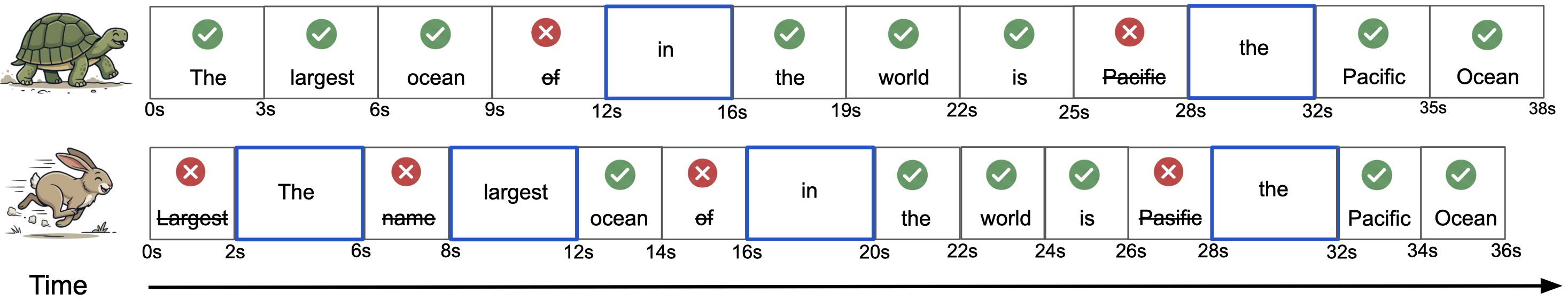}\vspace{-4mm}
    \caption{Example speculative-decoding traces for the same output sequence using draft models with different latencies and acceptance ratios (Accepted-to-draft token length ratio). The top row illustrates that a \textit{more accurate} but \textit{slower} draft model can result in \textit{longer overall decoding} time, highlighting the \textit{trade-off} between draft-model latency and acceptance ratio. }\vspace{-4.5mm}
    \label{fig:trade-off-intro}
    \vspace{-4.5mm}
\end{figure}

However, the efficiency of speculative decoding is inherently dependent on \textit{how fast and accurately} the draft model can generate the draft sequence. If the draft model has low latency but frequently proposes incorrect tokens, most of its predictions will be rejected by the verifier, and both the draft and verifier models will need to be invoked several times over to complete a sequence, resulting in slower text generation. Conversely, a highly accurate but computationally expensive draft model may offset the efficiency gains from integrating the smaller draft model into the decoding process (we visualize this trade-off in Figure \ref{fig:trade-off-intro}). Thus, a \textit{low latency architecture and target-draft agreement} are both important considerations for effective draft-model construction.

\looseness-1 Existing approaches to draft-model construction often treat architecture design and target-model alignment as separate stages. Some methods utilize an existing, independently trained small language model and adapt it to the target model through additional training or distillation~\citep{leviathan2023fast, hu2026adaspec}. However, aligning a separately trained small model with a large target model is challenging because independently trained models have distinct output distributions, making post-hoc alignment inherently difficult~\citep{li2025small}. To address the challenge of aligning the draft model's output distribution with the distribution of the target model, recent work has explored the idea of constructing sparse draft models by combining weight-level \textit{unstructured} pruning~\citep{sun2024simple, pmlr-v202-frantar23a} of the target model with self-distillation, while largely preserving the original architecture~\citep{lasby2025sd, monjur2026revisiting}. Here, less important weights are masked to produce a sparse model while preserving the original architecture. This approach can lead to better target-draft alignment because the draft model is derived directly from the \textit{target} model itself. However, unstructured pruning cannot provide latency benefits as irregular weight sparsity is difficult to exploit efficiently with conventional dense matrix-multiplication kernels in current Graphics Processing Unit (GPU) hardware~\citep{rath2026weight}. Meaningful latency improvements instead require \textit{structured pruning}, seeking to remove entire architectural components, such as layers or contiguous weight blocks~\citep{men2025shortgpt}. 

More recent work therefore proposes constructing draft models by structurally pruning the target model toward a manually chosen architecture to obtain a sufficiently small draft model~\citep{wei2026sdfp,yan2025decoding}. Despite their differences, these methods share a common limitation: they \textit{manually specify} the draft architecture before considering the trade-off between decoding efficiency and alignment with the target model. As a result, the resulting draft model is not explicitly optimized for speculative decoding performance. In practice, a design that achieves strong alignment may be unnecessarily expensive in terms of inference latency, while a highly efficient design may suffer from low alignment with the target model. Consequently, the resulting draft model may be far from the Pareto-optimal balance between latency and target-model agreement for a given target model and deployment environment. Designing effective draft models therefore requires \textit{jointly optimizing draft-model efficiency and target-draft agreement}, underpinning the basis of our work in this paper.

\textbf{Contributions.} To address this gap, we propose \texttt{T2D} (\textit{Target-to-Draft}), a self-distillation framework that jointly optimizes draft-model architecture and target-model alignment. As opposed to selecting a draft architecture beforehand and then adapting it to the target model, \texttt{T2D} learns a compact draft model through target-model pruning while distilling the target model's knowledge simultaneously. During training, both decoding latency and alignment with the target model directly influence the pruning decisions, allowing the architecture to emerge from the optimization process itself naturally. This enables \texttt{T2D} to automatically discover draft models that attain better speculative decoding performance, thereby balancing efficiency and acceptance rate. As we will show in subsequent sections, \texttt{T2D} can naturally converge to the best architecture among multiple candidate pruned architectures, compared to existing pruning-based draft-model construction baselines, including ShortGPT-style pruning~\citep{men2025shortgpt} and a fixed-architecture joint pruning-and-distillation baseline inspired by Sheared-LLaMA~\citep{xia2024sheared,yan2025decoding}. Across target LLMs from model families of varying sizes \texttt{T2D} can prune up to 92\% of the target model to produce highly efficient draft models. Finally, we show that \texttt{T2D} can also be utilized in a plug-and-play fashion with newer distillation methods such as AdaSpec~\citep{hu2026adaspec}, further underscoring its benefits. We summarize these contributions below:\vspace{-1mm}

\begin{itemize}[nosep, leftmargin=*]
\item We propose \texttt{T2D}, a self-distillation framework for speculative decoding that jointly optimizes draft-model architecture and target-draft alignment, allowing the draft architecture to emerge from the optimization process rather than being manually specified as in prior work.
\item To enable this joint optimization, we introduce a latency-aware structured pruning objective that directly incorporates the inference cost of candidate draft architectures, helping \texttt{T2D} effectively navigate decoding efficiency and alignment between the draft and target models.
\item We evaluate \texttt{T2D} across multiple LLM families and sizes (Llama-2, Vicuna-v1.5, Qwen2.5, Qwen3, and Nemotron), and find that it significantly outperforms heuristic architecture choices and pruning-based baselines, achieving up to \textbf{2.46}$\times$ in multilingual understanding domain and up to \textbf{1.89}$\times$ overall speculative decoding speedup across 11 diverse domains, such as coding, mathematics, reasoning, and multilingual understanding.
\end{itemize}

\section{Related Works}
\label{sec:related}

\textbf{Speculative Decoding.}
Speculative decoding reduces the inference cost of autoregressive decoding by predicting multiple future tokens in advance. \citet{stern2018blockwise}
first explored this idea using auxiliary output heads that predict future
tokens, which are verified by the same model under greedy decoding.
\citet{leviathan2023fast} and \citet{chen2023accelerating} generalized
the approach to sampling with a separate draft model and a rejection-sampling
acceptance rule that preserves the target model's output distribution.
Subsequent work has explored different algorithms for drafting and
verification. For example, Medusa~\citep{cai2024medusa} uses multiple decoding heads
and tree-structured candidates, EAGLE~\citep{li2024eagle} uses target-model
hidden features with a lightweight trainable module, and
SpecInfer~\citep{miao2023specinfer} verifies tree-structured candidates
from multiple drafts in a single pass. These methods primarily change the
drafting or verification algorithm. However, to successfully apply these algorithms,
a highly efficient draft model is very important. Thus, in this work, we focus on constructing high-performance draft models.

\textbf{Draft Model Construction.}
\looseness-1 A common approach to constructing draft models is to utilize separately pretrained language models, which can then be further trained to better match a target model~\citep{leviathan2023fast,hu2026adaspec}. More recent approaches derive the draft from the target itself. SD$^2$~\citep{lasby2025sd}, for example, combines target-model unstructured pruning with self-distillation and shows that this can improve draft-target alignment. However, unstructured pruning cannot reduce model size. Thus, structural approaches have also received attention because they produce smaller and computationally efficient dense models. \citet{wei2026sdfp} remove Transformer layers based on their importance, while \citet{yan2025decoding} investigate alternative depth-width configurations and show that model shape substantially affects speculative-decoding throughput. These approaches establish that different structural choices can lead to different decoding performance, motivating our work, which searches over such choices rather than fixing them beforehand.\vspace{-2mm}

\paragraph{Structured Pruning.}
Structured pruning algorithms reduce model size by removing layers, attention heads, or feed-forward channels by measuring the importance of Transformer blocks~\citep{ling2024slimgpt,men2025shortgpt, ashkboos2024slicegpt,ma2023llm,xia2024sheared, guo2025slimllm}. It provides a natural way to construct compact dense models with low latency for speculative decoding. Sheared-LLaMA~\citep{xia2024sheared} uses learnable pruning gates to obtain smaller models while satisfying predefined architectural constraints. However, these structured pruning approaches mostly focus on standard response-quality benchmarks such as MMLU~\citep{hendrycks2020measuring}, GSM8K~\citep{cobbe2021training}, SciQ~\citep{welbl2017crowdsourcing}, PIQA~\citep{bisk2020piqa}, WinoGrande~\citep{sakaguchi2021winogrande}, ARC-E~\citep{clark2018think}, and HellaSwag~\citep{zellers2019hellaswag}. To the best of our knowledge, no prior work has explored speculative-decoding-aware pruning, where model pruning is treated as a learnable objective for improving speculative decoding performance.

\section{Preliminaries and Background}
\label{sec:preli}

\looseness-1\textbf{LLMs and Autoregressive Decoding.}
An LLM $M_{\theta}$ generates an output sequence
$y$ autoregressively, producing one token at a time.
At decoding step $t$, the model predicts the next token, $y_t$ based on the input
prompt $x$ and the previously generated prefix
$y_{<t}=(y_1,\ldots,y_{t-1})$. Formally, it gives a conditional probability distribution for next-token prediction: $p_{\theta}(y_t \mid x, y_{<t})$.
LLMs are typically built from multiple stacked Transformer~\citep{vaswani2017attention} layers $L$, which defines the depth of the model. In
addition to depth, their computational cost is determined by several key
architectural dimensions, including the hidden dimension $d$, the number of
attention heads $H$, and the feed-forward intermediate dimension $I$.
These dimensions determine both the representational capacity of
the model and the cost of each decoding step.

In general,
larger models require more computation. For example, if
$\mathcal{F}_{\mathrm{lat}}(L,H,d,I)$ denotes the wall-clock latency of a
single decoding step for a model with architecture $(L,H,d,I)$, then $
\mathcal{F}_{\mathrm{lat}}(L,H,d,I)
\ge
\mathcal{F}_{\mathrm{lat}}(L',H,d,I)$, for $L \ge L'$, where $\mathcal{F}_{\mathrm{lat}}(.)$ is a latency estimator. A similar trend holds for other architectural dimensions, but less prominently \citep{yan2025decoding}. Note that with a smaller architecture, LLMs can be computationally much more efficient, motivating the research domain of speculative decoding, which we discuss next. 

\textbf{Speculative Decoding.}
Speculative decoding reduces autoregressive decoding latency by pairing a large
target model $M_{\theta_T}$ with a smaller draft model $M_{\theta_D}$. Given an input
$x$ and an already-generated and accepted prefix $\tilde{y}_{<t}$, the draft model autoregressively
proposes $\gamma$ candidate tokens $\hat{y}_{t:t+\gamma-1} = (\hat{y}_t,\hat{y}_{t+1},\ldots,\hat{y}_{t+\gamma-1})$.
The target model then verifies all $\gamma$ candidates in a single forward pass by matching the next-token prediction of the target model and the proposed token of the draft model at each position. A draft token is accepted if it matches the target model's own
prediction at that position. However, only the initial consecutive accepted tokens are
kept. If the first mismatch occurs after $a$ accepted tokens, the remaining draft
tokens are discarded, and the target model's prediction at the mismatch position is
used in their place. Thus, each verification round extends the generated
sequence by $a+1$ tokens.

The resulting speedup of speculative decoding depends primarily on the average number of accepted draft
tokens, the draft-model latency, the target-model verification latency, and the
draft length $\gamma$. Let $\mathcal{T}_D$ and $\mathcal{T}_T$ denote the per-token decoding latencies
of the draft and target models, and let the latency ratio, $c = \mathcal{T}_D / \mathcal{T}_T$. Under the standard
approximation that a target verification pass latency is close to the target
decoding step latency, after simplification, the speculative-decoding speedup or ratio of speculative-decoding and autoregressive decoding throughput can be written as:
$
\mathcal{S} = \frac{\mathbb{E}[a]+1}{(\gamma c + 1)},
$
where $\mathbb{E}[a]$ is the average accepted tokens~\citep{leviathan2023fast} and $\mathbb{E}[a]+1$ is the average accepted length including the correction token provided by the target model. For a fixed draft length,
this expression shows that speculative decoding benefits from a draft model that
is both \textit{computationally efficient} and \textit{well aligned} with the target: lowering $c$
shrinks the denominator, while better draft-target alignment increases $\mathbb{E}[a]$.

Notably, these two objectives are not independent, as improving one often comes at the expense of the other during draft model design. Knowledge distillation (KD)~\citep{xu2024survey} is an effective approach for improving draft-target alignment, while structured pruning of the target model~\citep{lasby2025sd, xia2024sheared, yan2025decoding} is a promising direction for reducing draft-model latency while preserving alignment. In the following sections, we discuss these approaches in detail.

\textbf{Knowledge Distillation for Draft-Model Alignment.}
\looseness-1 Knowledge distillation improves draft-target alignment by training the draft
model to reproduce the target model's output distribution. One way to perform distillation is to
directly minimize the KL divergence between the two next-token distributions~\citep{zhou2024distillspec},
$D_{\mathrm{KL}}\big(p_{\theta_T}(\cdot\mid x,y_{<t}) \,\|\, p_{\theta_D}(\cdot\mid x,y_{<t})\big)$.
Alternatively, training sequences, $\tilde{y} = (\tilde{y}_1, \ldots, \tilde{y}_N)$, can be generated directly from the target model and can then be used to train the draft
model with a standard causal cross-entropy (CE) loss:\vspace{-5mm}

{\small
\begin{equation}
\mathcal{L}_{\mathrm{CE}}(\theta_D, \tilde{y})
=
-\frac{1}{N} \sum_{t=1}^{N} \log p_{\theta_D}(\tilde{y}_t \mid x, \tilde{y}_{<t}).
\label{eq:kd_ce}
\end{equation}\vspace{-3mm}
}


\textbf{Structured Pruning.}
Structured pruning removes entire architectural components of an LLM, such as Transformer layers, attention heads, or feed-forward channels, rather than individual weights. Given
an LLM with dimensions $(L,H,d,I)$, structured pruning yields a smaller \textit{dense}
model with dimensions $(L',H',d',I')$, where each retained dimension is smaller
than the original.
\looseness-1 A structured-pruning pipeline generally requires a fixed architecture to be specified beforehand~\citep{xia2024sheared, men2025shortgpt,yan2025decoding}. However, for speculative decoding, we do not want to determine the architecture heuristically. Rather, we want to prune the target model as much as possible while maintaining high alignment between the pruned model and the target model. This forms the basis for our technical contributions in this paper. To achieve this aim, we introduce \texttt{T2D}, a joint \textit{pruning-and-alignment} framework that learns the trade-off between model efficiency and target-model alignment to construct an effective draft model for speculative decoding. We describe our proposed \texttt{T2D} framework in the next section.

{
\begin{algorithm}[t]
\fontsize{9}{9}\selectfont
\caption{\texttt{T2D}: Target-to-Draft Model Construction}
\label{alg:full-pipeline}
\begin{algorithmic}[1]
\Statex \textbf{Input:} Target model $M_{\theta_T}$, target model latency $\mathcal{T}_T$, alignment data $\mathcal{D}$, frozen latency predictor $\mathrm{MLP}_{\mathrm{lat}}$, reference CE $\mathcal{L'}_{\mathrm{CE}}$, number of epochs $Ep$.
\Statex \textbf{Output:} Dense draft model $M_{\theta_D^*}$

\State Initialize gates $\{\alpha_j\}$, $\lambda_1\leftarrow0$, $\lambda_2\leftarrow0$, $\widehat{\mathcal{L}}_{\mathrm{CE}}\leftarrow0$
\For {$e=1,\ldots, Ep$}
\For{$s=1,\ldots,S$}
    \State Sample Hard-Concrete gates and compute expected-open probabilities $\{\bar z_{\ell}^{L},\bar z_h^{H},\bar z_{\ell,i}^{I}\}$
    \State Compute estimated continuous architecture $(\Lest,\Hest,\Iest,\dest)$ via Eq.~\ref{eq:estimated-dims}
    \State $\bar{\mathcal{T}}\leftarrow\exp\!\big(\mathrm{MLP}_{\mathrm{lat}}(\Lest,\Hest,\Iest)\big)$
    \State $\mathcal{L}_{\mathrm{lat}}\leftarrow\bar{\mathcal{T}}/\mathcal{T}_T$
    \State Apply sampled gates to $M_{\theta_T}$ and compute $\mathcal{L}_{\mathrm{CE}}$ using target-generated tokens $\tilde y$ from $\mathcal{D}$ \hfill \Comment{Eq.~\ref{eq:kd_ce}}
    \State $\widehat{\mathcal{L}}_{\mathrm{CE}}\leftarrow(1-m)\widehat{\mathcal{L}}_{\mathrm{CE}}+m\mathcal{L}_{\mathrm{CE}}$
    \State $g \leftarrow \max\!\left(\dfrac{\widehat{\mathcal{L}}_{\mathrm{CE}} - \mathcal{L'}_{\mathrm{CE}}}{\widehat{\mathcal{L}}_{\mathrm{CE}}},\, 0\right)$
    \State Calculate loss, $\mathcal{L}$ via Eq.~\ref{eq:loss}.
    \State Update model and gate parameters by gradient descent on $\mathcal{L}$
    \State $\lambda_1\leftarrow\operatorname{clip}(\lambda_1+\eta_1g,\,0,\,\lambda_1^{\max})$,\quad
    $\lambda_2\leftarrow\operatorname{clip}(\lambda_2+\eta_2g,\,0,\,\lambda_2^{\max})$
\EndFor
\EndFor
\State Get final continuous values $\bar L_D^*, \bar H_D^*, \bar I_D^*$
\State Select retained layers $L_D^*$, attention heads $H_D^*$, and FFN channels $I_D^*$ from the optimized gates
\State Instantiate a dense architecture with $d_D^*=H_D^*d_{\mathrm{head}}$ 
\State Copy retained parameters from $M_{\theta_T}$
\State Continually fine-tune the obtained model on $\mathcal{D}$ for one epoch to obtain a dense and aligned $M_{\theta_D^*}$
\State \Return $M_{\theta_D^*}$
\end{algorithmic}
\end{algorithm}
}

\section{Proposed Method: \texttt{T2D}}
\label{sec:method}

In Section~\ref{sec:preli}, we highlighted a key challenge in draft-model construction~\citep{lasby2025sd}: a smaller draft model reduces decoding latency but may also lead to lower token acceptance during speculative decoding.
Thus, given a target model $M_{\theta_T}$, our goal is to construct a compact draft
model by selecting an architecture that balances decoding latency and
draft-target alignment. Let $(L_D,H_D,d_D,I_D)$ denote the architectural dimensions
of the draft model. We formulate draft-model construction as:\vspace{-6mm}

\begin{equation}
\resizebox{0.92\linewidth}{!}{$
(L_D^*,H_D^*,d_D^*,I_D^*)
=
\arg\min_{(L_D,H_D,d_D,I_D)\in\mathcal{A}}
\Big(
\mathcal{F}_{\mathrm{lat}}(L_D,H_D,d_D,I_D),
\mathcal{L}_{\mathrm{CE}}(L_D,H_D,d_D,I_D;\mathcal{D})
\Big),
$}
\label{eq:problem}
\end{equation}\vspace{-1mm}

where $\mathcal{A}$ denotes the set of architectures reachable through structured pruning, and $\mathcal{D}=\{(x_1,\tilde{y}_1),(x_2,\tilde{y}_2),\ldots,(x_S,\tilde{y}_S)\}$ denotes the set of query-response pairs generated by $M_{\theta_T}$ and used for optimization.
Directly optimizing Eq.~\ref{eq:problem} is difficult because the architectural variables are discrete and the search space is combinatorial. We therefore introduce \textit{differentiable gates} that convert discrete pruning decisions into continuous variables, enabling gradient-based optimization inspired by~\citet{xia2024sheared}. Similarly, model latency cannot be incorporated directly into the optimization loss since $\mathcal{F}_{\mathrm{lat}}(\cdot)$ is not differentiable by definition, necessitating a differentiable proxy. In the following sections, we describe how the gates represent a differentiable architecture, how we obtain a differentiable latency estimate, and how we formulate the loss function for joint pruning and KD-based alignment training. Additionally, we present the overall \texttt{T2D} framework as Algorithm~\ref{alg:full-pipeline}.

\textbf{Learnable Structured Pruning.}
\label{subsec:learnable-pruning} To enable gradient-based optimization, we introduce learnable gates over removable Transformer structures,
including layer gates $z_{\ell}^{L}$, attention-head gates $z_h^{H}$,
and feed-forward channel gates $z_{l,i}^{I}$.
\looseness-1 For each gate parameter
$\alpha_j$, we adopt the Hard-Concrete
relaxation~\citep{louizos2017learning} and compute an expected-open probability
$\bar z_j \in [0,1]$, which serves as a differentiable approximation of a
binary keep-or-remove decision (Lines 1-4 in Algorithm \ref{alg:full-pipeline}). Using these probabilities, we estimate the architecture dimensions as:

{
\begin{equation}
\Lest = \sum_{\ell}\bar z_{\ell}^{L},
\qquad
\Hest = Q\sum_h\bar z_h^{H},
\qquad
\Iest =
\frac{
\sum_{\ell}\bar z_{\ell}^{L}
\sum_i\bar z_{\ell,i}^{I}
}
{
\sum_{\ell}\bar z_{\ell}^{L}
},
\label{eq:estimated-dims}
\end{equation}
}\vspace{-5mm}

where $Q$ denotes the number of query heads associated with each key-value
group (Line 5). We
estimate the hidden dimension as $\dest = \Hest d_{\mathrm{head}}$, where $d_{\mathrm{head}}$ is the head dimension. Now that we have formalized the learnable gates, we will relate latency to the learnable gates to obtain differentiable latency.

\textbf{Differentiable Latency Modeling}
\label{subsec:latency-predictor} Measuring hardware latency directly for every candidate architecture
during optimization is both expensive and, more importantly,
non-differentiable. To estimate \textit{differentiable latency} directly from $(\Iest,\Lest,\Hest)$, we train a lightweight latency predictor \textit{offline}, and keep it frozen while optimizing the pruning gates.
\looseness-1To train the latency predictor, for each candidate dense architecture, $(L_D,H_D,I_D)$, we measure its single-token decoding
latency, $\mathcal{T}_{dec}$, and generate a latency-measurement dataset. Then we train a small regression model $\mathrm{MLP}_{\mathrm{lat}}$ with the dataset. The latency predictor $\mathrm{MLP}_{\mathrm{lat}}$
takes the architectural dimensions $(L_D,H_D,I_D)$ as input and predicts the logarithm
of the corresponding latency. Further description of the training of the latency predictor is provided in Appendix \ref{app:mlp-lat-pred}.

\looseness-1During gate optimization, we pass $(\Lest,\Hest,\Iest)$ through the frozen
latency predictor. We obtain the predicted latency as, ${\bar{\mathcal{T}}} = \exp\big(\mathrm{MLP}_{\mathrm{lat}}(\Lest,\Hest,\Iest)\big)$ (Line 6).
Because this predictor is differentiable with respect to its architectural inputs,
gradients can flow from the latency objective all the way back to the pruning
gates during backpropagation, giving us a latency-aware training signal. With learnable
latency via $\bar{\mathcal{T}}$ and alignment via Eq.~\ref{eq:kd_ce}, we now combine them into the
single joint objective that drives our pruning procedure.

\begin{wrapfigure}[23]{r}{0.45\textwidth}
    \centering\vspace{-7.5mm}
    \includegraphics[width=0.45\textwidth]{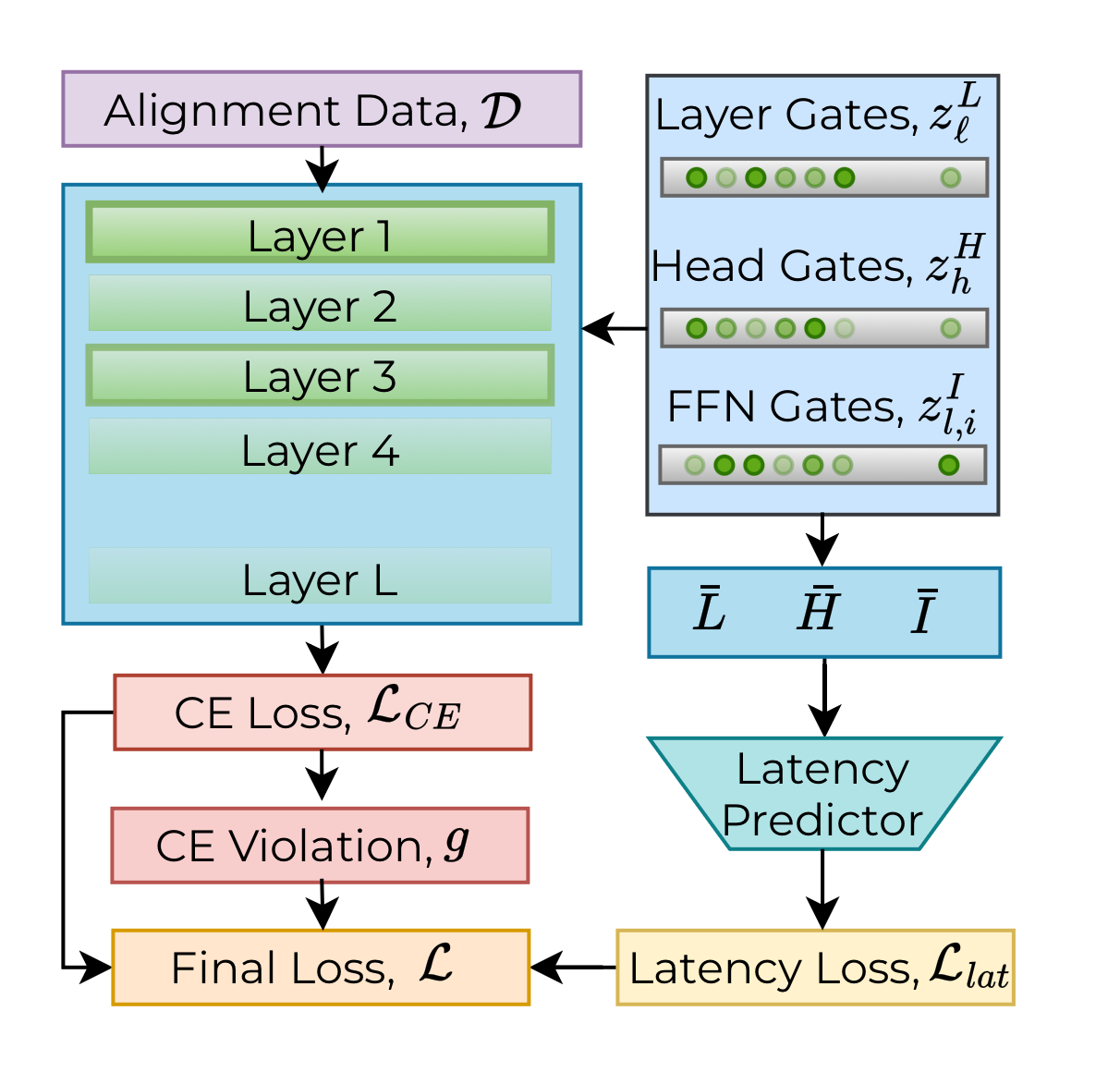}\vspace{-9mm}
    \caption{Overview of the \texttt{T2D} training pipeline. Learnable gates mask layers, attention heads, and FFN channels to jointly optimize the pruned architecture and model weights using a frozen latency predictor, target-generated cross-entropy loss, and a reference-guided penalty.}
    \label{fig:full_pipeline}
\end{wrapfigure}

\textbf{Latency-Aware Pruning and Knowledge Distillation}
\label{subsec:mask-objective} We optimize the gates using three complementary loss terms: (1) a latency objective, (2) the CE loss, and (3) a reference-guided penalty that limits how much the CE is allowed to degrade. Our entire framework is visualized in Figure \ref{fig:full_pipeline}.
The latency objective normalizes the predicted draft latency by the measured
latency of the unpruned target model: $\mathcal{L}_{\mathrm{lat}} = \bar{\mathcal{T}} / \mathcal{T}_T$ (Line 7).
The raw CE loss $\mathcal{L}_{\mathrm{CE}}$ is computed on the currently masked
model using the target-generated response $\tilde y$ from the alignment data by using Eq.~\ref{eq:kd_ce}.
(Line 8). We weight the latency and CE terms by $\mu_{\mathrm{lat}}$
and $\mu_{\mathrm{CE}}$, respectively.

Since optimization starts from the \textit{unpruned} target model, the latency term needs to be dominant during the early stages of training to initiate pruning. However, if pruning is too aggressive, the CE loss can spike rapidly as useful structures are masked too quickly. To guard against this, we impose a reference-based one-sided augmented-Lagrangian penalty. The penalty remains inactive while the smoothed CE, $\widehat{\mathcal{L}}_{\mathrm{CE}}$, remains below the reference $\mathcal{L'}_{\mathrm{CE}}$, and increases only when the constraint is violated. We calculate the smoothed CE loss, $\widehat{\mathcal{L}}_{\mathrm{CE}}$, using a smoothing factor $m$, which controls how much the current loss influences the penalty (Line 9).
Thus, our final optimization loss becomes:\vspace{-4mm}

\begin{equation}
    \mathcal{L}=
\mu_{\mathrm{lat}}\mathcal{L}_{\mathrm{lat}}
+\mu_{\mathrm{CE}}\mathcal{L}_{\mathrm{CE}}
+\lambda_1 g+\lambda_2 g^2,
\label{eq:loss}
\end{equation} 

where $g$ determines the amount of constraint violation, calculated similarly to Line 10 of Algorithm~\ref{alg:full-pipeline}, and $\lambda_1$ and $\lambda_2$ are the learnable penalty weights, whose learning is controlled by the hyperparameters $\eta_1$, $\eta_2$, $\lambda_1^{\mathrm{max}}$, and $\lambda_2^{\mathrm{max}}$ (Line 13). Finally, we update the model weights, gates, and penalty weights based on the calculated loss (Lines 11-13). This joint objective pushes the estimated decoding latency down while actively preventing alignment from drifting significantly from the acceptable tolerance. The two goals motivated in Equation~\ref{eq:problem} of Section~\ref{sec:preli} are therefore optimized \textit{simultaneously}, over the \textit{same} gate parameters, rather than in separate stages. After optimization, we convert the learned continuous gate scores into a concrete draft architecture, as described next.

\looseness-1\textbf{Draft Model Construction and Continual Finetuning.}
\label{subsec:materialization} We convert the continuous gate scores into discrete decisions after the joint training (Lines 16-18). We then instantiate a new, dense Transformer using the selected layers, attention structures, and feed-forward channels, and copy their corresponding parameters from the trained target model (Line 19). Finally, we fine-tune the resulting draft model on $\mathcal{D}$ for one epoch to let the weights readjust and mitigate the effects of converting continuous gate values into discrete decisions. Thus, we obtain the final draft model $M_{\theta_D^*}$, derived from $M_{\theta_T}$ by truly reducing the number of parameters.


\section{Experiments and Results}
\label{sec:results}

Having described \texttt{T2D} in Section~\ref{sec:method}, we now
explain our experimental setup, including the target
and draft models, the training and validation datasets, benchmark dataset, and evaluation metrics. We then discuss our results subsequently.

 \textbf{Models.} We evaluate \texttt{T2D} on Llama-2-7B-Chat-HF~\citep{touvron2023llama}, which has been widely used as a base model for both pruning and speculative decoding research~\citep{yan2025decoding,sun2024simple}, enabling direct comparisons with prior work and its design decisions. To assess the scalability and adaptability of \texttt{T2D} across model sizes and families, we further evaluate it on Vicuna-13B-v1.5~\citep{zheng2023judging}, Qwen2.5-7B-Instruct~\citep{hui2024qwen2}, Qwen3-14B~\citep{yang2025qwen3}, and Nemotron-OpenReasoning-32B (Nemotron-32B)~\citep{ahmad2025opencodereasoning}. Together, these models provide a representative evaluation across different model families and scales. Additionally, Qwen3-14B and Nemotron-OpenReasoning-32B are reasoning models, enabling us to evaluate \texttt{T2D} on both reasoning and non-reasoning response types. For each model family, we construct the draft model from the corresponding target model, following the procedure described in Section~\ref{sec:method}, and use it for speculative decoding.

\textbf{Training Dataset Construction.} For each target model, we construct a target-generated alignment
dataset inspired by~\citet{wu2024lamini} from a mixture of publicly available instruction tuning corpora,
including Dolly-15K~\citep{conover2023free}, Alpaca~\citep{taori2023stanford}, WizardLM~\citep{xu2024wizardlm}, UltraChat~\citep{ding2023enhancing}, OpenOrca~\citep{lian2023openorca}, WildChat~\citep{zhao2024wildchat}, MetaMathQA~\citep{yu2024metamath}, CodeFeedback~\citep{zheng2024opencodeinterpreter},
Bactrian-X~\citep{li2023bactrian}, and xP3~\citep{muennighoff2023crosslingual}.
We construct 250,000 question-answer pairs for the training set $\mathcal{D}$, which is used during pruning and alignment optimization, and randomly sample 5,000 pairs for the validation set $\mathcal{D}_{\mathrm{val}}$. The validation set is used only for hyperparameter optimization. We provide further details of the training dataset in Appendix \ref{app:training-dataset}.

\textbf{Benchmark Dataset.} We evaluate the performance of the draft models on the SPEED Bench~\citep{abramovich2026speed} speculative decoding benchmark, which contains 798 queries across 11 domains. We provide details of the SPEED benchmark in Appendix~\ref{app:speed}.

\textbf{Baselines.} Given that target-model-pruning-based draft-model construction is an emerging research direction with limited prior work, we compare \texttt{T2D} against state-of-the-art draft-construction approaches that isolate the effect of joint architecture optimization: \textit{Wide} Draft Models~\citep{yan2025decoding} and ShortGPT~\citep{men2025shortgpt}. Additional details of these baselines are provided in Appendix~\ref{app:imp-details} due to space limits. Similar to prior work~\citep{leviathan2023fast}, we report the speculative decoding speedup, $\mathcal{S}$,
relative to autoregressive decoding as our main metric in results.

Now, we discuss and compare \texttt{T2D} with the baselines. We first evaluate \texttt{T2D} 's performance under the simple distillation process described in Section~\ref{sec:method}. We then analyze the trade-off curve that \texttt{T2D} seeks to optimize and examine how it balances latency and CE loss during joint optimization. Next, we show that \texttt{T2D}-derived draft models can be utilized in a plug-and-play fashion with speculative decoding-specific alignment techniques. Finally, we compare \texttt{T2D} with independently developed, extensively trained draft models and show that such models are not well suited for speculative decoding because they are optimized only for better text generation, not explicitly for this objective.

\begin{table*}[t]
\centering
\caption{Speculative decoding speedup across different evaluation domains. 
For each target model, we compare draft models constructed by ShortGPT, \textit{Wide}, and \texttt{T2D}. 
The highest speedup ($\times$) for each target model and domain is shown in \textbf{bold}.}\vspace{5mm}
\label{tab:speedup}
\fontsize{8}{9}\selectfont
\setlength{\tabcolsep}{3.5pt}
\resizebox{\textwidth}{!}{
\begin{tabular}{llcccccccccccc}
\toprule
\textbf{Target Model} & \textbf{Draft}
& \makecell{\textbf{Code}}
& \makecell{\textbf{Hum.}}
& \makecell{\textbf{Math}}
& \makecell{\textbf{Multilin.}}
& \makecell{\textbf{QA}}
& \makecell{\textbf{RAG}}
& \makecell{\textbf{Reason.}}
& \makecell{\textbf{Role.}}
& \makecell{\textbf{Stem}}
& \makecell{\textbf{Summ.}}
& \makecell{\textbf{Writing}}
& \makecell{\textbf{Overall}} \\
\midrule



\multirow{3}{*}{Llama-2-7B-Chat}
& ShortGPT
& 1.72 & 1.30 & 1.57 & 1.46 & 1.06 & 1.57 & 1.66
& 1.20 & 1.43 & 1.41 & 1.41 & 1.45 \\
& \textit{Wide}
& 1.82 & 1.46 & 1.74 & 1.83 & 1.34 & 1.63 & 1.73
& 1.25 & 1.58 & 1.43 & 1.50 & 1.59 \\
& \texttt{T2D}
& \textbf{1.98} & \textbf{1.59} & \textbf{1.92} & \textbf{2.03} & \textbf{1.46} & \textbf{1.79} & \textbf{1.90} 
& \textbf{1.35} & \textbf{1.72} & \textbf{1.60} & \textbf{1.62} & \textbf{1.74} \\

\midrule

\multirow{3}{*}{Vicuna-13B-v1.5}
& ShortGPT
& 1.95 & 1.66 & 1.82 & 2.22 & 1.40 & 1.74 & 1.86
& 1.30 & 1.70 & 1.73 & 1.62 & 1.74 \\
& \textit{Wide}
& 2.06 & 1.67 & 1.92 & 2.22 & 1.44 & 1.79 & 1.91
& 1.33 & 1.72 & 1.74 & 1.66 & 1.78 \\
& \texttt{T2D}
& \textbf{2.13} & \textbf{1.79} & \textbf{1.99} & \textbf{2.46} & \textbf{1.47} & \textbf{1.97} & \textbf{2.04}
& \textbf{1.39} & \textbf{1.83} & \textbf{1.88} & \textbf{1.73} & \textbf{1.89} \\

\midrule

\multirow{3}{*}{Qwen2.5-7B-Instruct}
& ShortGPT
& 1.50 & 1.12 & 1.50 & 1.43 & 1.00 & 1.20 & 1.35
& 0.88 & 1.24 & 1.07 & 1.15 & 1.24 \\
& \textit{Wide}
& 1.59 & 1.15 & 1.56 & 1.43 & 1.04 & 1.22 & 1.42
& 0.90 & 1.29 & 1.05 & 1.19 & 1.28 \\
& \texttt{T2D}
& \textbf{1.73} & \textbf{1.32} & \textbf{1.71} & \textbf{1.65} & \textbf{1.14} & \textbf{1.39} & \textbf{1.54}
& \textbf{1.02} & \textbf{1.44} & \textbf{1.22} & \textbf{1.29} & \textbf{1.42} \\

\midrule

\multirow{3}{*}{Qwen3-14B}
& ShortGPT
& 1.09 & 1.06 & 1.13 & 1.16 & 1.03 & 1.12 & 1.12
& 0.87 & 1.11 & 0.99 & 0.96 & 1.07 \\
& \textit{Wide}
& 1.13 & 1.09 & 1.17 & 1.20 & 1.07 & 1.15 & 1.16
& 0.90 & 1.14 & 1.00 & 0.99 & 1.10 \\
& \texttt{T2D}
& \textbf{1.23} & \textbf{1.21} & \textbf{1.27} & \textbf{1.31} & \textbf{1.17} & \textbf{1.26} & \textbf{1.26}
& \textbf{1.00} & \textbf{1.25} & \textbf{1.12} & \textbf{1.09} & \textbf{1.20} \\

\midrule

\multirow{3}{*}{Nemotron-32B}
& ShortGPT
& 1.51 & 1.33 & 1.49 & 1.56 & 1.41 & 1.61 & 1.47
& 1.27 & 1.34 & 1.46 & 1.29 & 1.43 \\
& \textit{Wide}
& 1.82 & 1.55 & 1.82 & 1.94 & 1.64 & 1.92 & 1.76
& 1.47 & 1.58 & 1.76 & 1.55 & 1.72 \\
& \texttt{T2D}
& \textbf{1.87} & \textbf{1.64} & \textbf{1.87} & \textbf{2.04} & \textbf{1.74} & \textbf{2.03} & \textbf{1.80}
& \textbf{1.55} & \textbf{1.66} & \textbf{1.84} & \textbf{1.61} & \textbf{1.79} \\

\bottomrule
\end{tabular}
}
\vspace{-9mm}
\end{table*}

\subsection{Main Results}
\looseness-1Table~\ref{tab:speedup} presents the decoding speedup achieved by all methods. \texttt{T2D} outperforms all baselines across all domains. For Llama-2-7B-Chat, the \texttt{T2D}-derived draft model achieves an overall speedup of up to 1.74$\times$, which is 20\% higher than ShortGPT and 9.4\% higher than the \textit{Wide} model. For the larger Vicuna-13B-v1.5 model, \texttt{T2D} achieves an overall speedup of 1.89$\times$, which is more than 6.2\% higher than the heuristic baselines, outperforming them across all domains and reaching a maximum speedup of 2.46$\times$ in the Multilingual category. Similar trends appear for the other models (Qwen2.5-7B-Instruct, Qwen3-14B, and Nemotron-32B), demonstrating that \texttt{T2D} can effectively find a better trade-off between architecture and alignment and achieve the best overall standard speculative decoding performance, even for reasoning models (Qwen3-14B and Nemotron-32B) that generate complex reasoning tokens. In contrast, the heuristic baselines sometimes achieve speedups below 1.00$\times$. For example, the draft models derived from Qwen2.5-7B-Instruct and Qwen3-14B achieve speedups of about 0.9$\times$ in the Roleplay category, indicating worse performance than autoregressive decoding. We provide additional details on the architectures derived by \texttt{T2D} in Appendix~\ref{app:archi}. Our qualitative analysis on Llama-2 shows that the \texttt{T2D} draft model generates mostly coherent responses even after 88\% pruning, demonstrating the effectiveness of joint optimization. Due to space constraints, we provide qualitative analysis in Appendix~\ref{app:sd-performance}.

\subsection{Why is \texttt{T2D} better than random manual architecture selection?}

\begin{wrapfigure}[14]{r}{0.35\textwidth}
    \centering \vspace{-7mm}
    \includegraphics[width=0.35\textwidth]{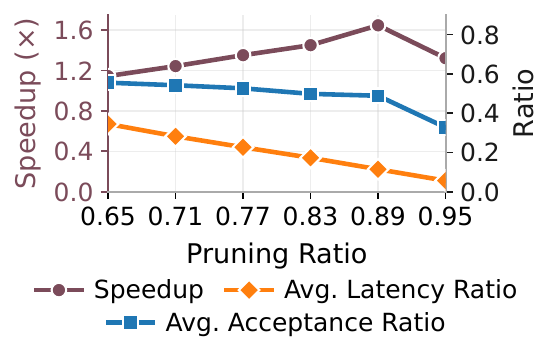}\vspace{-10mm}
    \caption{Trade-off curve of acceptance ratio and latency across pruning ratio and speedup of the draft model across pruning ratio of Llama-2-7B-Chat model when pruned with ShortGPT.}
    \label{fig:trade-off}
\end{wrapfigure}

\looseness-1 To investigate the reason behind the effectiveness of \texttt{T2D}, we performed a pruning ratio sweeping experiment in which we pruned the Llama-2-7B-chat-hf model by 65\%, 71\%, 77\%, 83\%, 89\%, and 95\% with ShortGPT. After pruning, we fine-tuned the resulting draft models for one epoch on target-model-generated responses. Finally, after fine-tuning, we evaluated the models on the SPEED Benchmark and plotted the change in speedup, the latency ratio relative to the target model, and the average acceptance ratio relative to the maximum acceptance length. Figure \ref{fig:trade-off} shows the trade-off between acceptance length and latency for different pruning ratios. With aggressive pruning, both the latency and the acceptance length of the draft model decrease. However, speedup peaks at 89\% pruning and then declines, showing that the best trade-off occurs around 89\% pruning, where the rate of latency decrease outweighs the rate of acceptance-rate decrease. As a result, speedup is highest at that point. \texttt{T2D} finds this maximum speedup point by pruning Llama-2-7B-Chat naturally up to 88\% and reducing latency while keeping the acceptance ratio decrease rate as low as possible. As a result, the \texttt{T2D}-developed draft model achieves the best overall speedup in an \textit{automated} fashion, whereas the other pruning baselines require a fixed prior pruning budget, limiting their efficacy.

\subsection{How does \texttt{T2D} prune the model while balancing CE loss and latency?}

\begin{wrapfigure}[15]{r}{0.4\textwidth}
\centering \vspace{-4mm}
\includegraphics[width=0.4\textwidth]{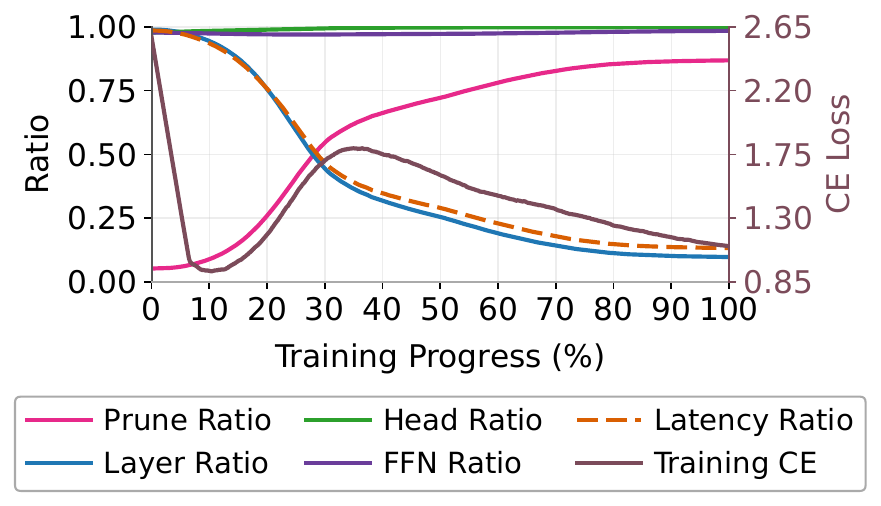}\vspace{-7mm}
\caption{Pruning ratio of the Llama-2-7B-Chat model over the training period along with the kept percentage (ratio) of the Layers, Heads, and FFN. We also plot latency and CE loss.}
\label{fig:trend-llama-7b}
\end{wrapfigure}

\looseness-1 In Figure \ref{fig:trend-llama-7b}, we show the ratios of different LLM components along with the pruning ratio for Llama-2-7B-Chat. Additionally, we plot the latency and the CE loss over the training period. Initially, the model has very high CE loss because the gates are randomly initialized, and very high latency because pruning has not yet started. After some steps, the CE loss goes down as the model weights adjust. Then, near 10\% of training, pruning starts after a warm-up period, and the CE loss begins to increase. After the CE loss rises to a certain value, the Lagrange penalty increases, and the penalty weights $\lambda_1$ and $\lambda_2$ are increased significantly. As a result, the CE loss starts to decrease after that point, and the pruning rate is balanced against the CE loss. After 88\% pruning, the pruning ratio flattens out. Over the training period, layer pruning contributes more, while heads and FFN are pruned less, because most of the latency benefit comes from layer pruning, suggesting that wider models are better for speculative decoding. We present the pruning ratio, latency, and CE loss curves for other models in Appendix \ref{app:trend-others}. The overall trends are similar to Llama-2-7B-Chat. However, for larger models, \texttt{T2D} prunes the heads and FFN as well, since these components have higher dimensions in larger models and also contribute a non-negligible portion of the latency. \texttt{T2D} achieves a better latency-alignment trade-off under simple supervised fine-tuning-based distillation. However, more advanced alignment techniques have been proposed specifically for speculative decoding, such as AdaSpec~\citep{hu2026adaspec}. In the next section, we investigate whether \texttt{T2D} continues to outperform alternative methods when combined with speculative decoding-specific alignment techniques.

\subsection{Can \texttt{T2D} Models Adapt to Other Alignment Methods?}

\begin{wraptable}[10]{r}{0.24\textwidth}
\centering\vspace{-8mm}
\caption{Speculative decoding performance comparison of draft models under the AdaSpec distillation method (best in \textbf{bold})}
\vspace{5pt}
\label{tab:adaspec-perf}
\fontsize{8}{9}\selectfont
\setlength{\tabcolsep}{3.5pt}
\begin{tabular}{lccc}
\toprule
\textbf{Draft Model} & \textbf{Speedup} \\
\midrule
ShortGPT & 1.43 \\
\textit{Wide} & 1.59 \\
\texttt{T2D} & \textbf{1.79} \\
\bottomrule
\end{tabular}
\end{wraptable}

In our experiments, we utilized target model-generated responses for distillation and fine-tuned the draft model accordingly. However, for even better alignment and performance, \texttt{T2D}-derived draft models can also be adapted to more advanced alignment methods such as AdaSpec. To evaluate whether \texttt{T2D} models can continue improving under domain-specific specialized alignment, we fine-tune all Llama-2-7B-Chat-based draft models using AdaSpec on the GSM8K dataset~\citep{cobbe2021training} and evaluate their speculative decoding performance on the GSM8K test set, following the same evaluation protocol as AdaSpec. In Table~\ref{tab:adaspec-perf}, we present the performance of all draft models. Under AdaSpec distillation on the GSM8K dataset, \texttt{T2D} outperforms the other baselines, achieving up to 20\% higher speedup. This shows that \texttt{T2D} is a superior draft-model construction algorithm that can be adapted to other distillation methods in a plug-and-play fashion and achieve even greater speedups than heuristic architecture-selection methods.\vspace{2mm}
%

\subsection{Our \texttt{T2D} draft models better than an independently trained draft model?}

\begin{wraptable}[10]{r}{0.35\textwidth}
\centering\vspace{-7mm}
\caption{Speculative decoding speedup ($\times$) comparison between finetuned Sheared-Llama and \texttt{T2D} with Llama-2-7B-Chat as the target model. The highest speedup is shown in \textbf{bold}.}
\vspace{4pt}
\label{tab:sheared-AL-speed}
\fontsize{8}{9}\selectfont
\setlength{\tabcolsep}{3.5pt}
\begin{tabular}{lccc}
\toprule
\textbf{Draft Model} & \textbf{Size (B)} & \textbf{Speedup} \\
\midrule
Sheared-Llama-1.3B-ft & 1.27  & 1.18 \\
\texttt{T2D} & 0.86 & \textbf{1.74} \\
\bottomrule
\end{tabular}
\end{wraptable}

We compare the \texttt{T2D}-produced Llama-2-7B-Chat draft model with the Sheared-Llama-1.3B model, which is an independently pretrained model on 50B tokens, which is about 340$\times$ more data than \texttt{T2D} takes during optimization. We fine-tune the Sheared-Llama-1.3B model with the Llama-2-7B-chat-hf generated responses similarly to \texttt{T2D} for one epoch to readjust the weights. We call the finetuned model Sheared-Llama-1.3B-ft and report the overall speedup of both models in Table \ref{tab:sheared-AL-speed}. The Sheared-Llama-1.3B-ft model achieves only a $1.18\times$ overall speedup, whereas the \texttt{T2D} draft model achieves a significantly higher $1.74\times$ speedup. This is because Sheared-Llama is designed for general text generation quality rather than optimizing the latency-alignment trade-off for speculative decoding. As a result, it delivers lower speculative decoding performance than \texttt{T2D}. These results further demonstrate that \texttt{T2D} is a more effective draft-model construction method for identifying speculative-decoding-aware draft models that can significantly accelerate autoregressive decoding. We provide per-domain speedup comparison in Appendix \ref{app:per-domain-sheared}.\vspace{2mm}

\section{Conclusion}
\label{sec:conclusion}

In this work, we address a key limitation of existing draft-model construction methods for speculative decoding: their reliance on fixed draft architectures chosen before considering alignment with the target model. We propose \texttt{T2D}, a self-distillation framework that jointly optimizes draft-model architecture and target-model alignment, allowing the draft architecture to emerge directly from the optimization process. To enable this, we introduce a latency-aware structured pruning objective that combines a differentiable estimate of inference latency with a target-distilled cross-entropy objective, guided by a reference-based penalty that prevents pruning from outpacing alignment. Experiments across the Llama-2, Vicuna-v1.5, Qwen2.5, Qwen3, and Nemotron families show that \texttt{T2D} consistently outperforms heuristic architecture choices and pruning-based baselines, demonstrating that effective draft models can be discovered directly from the target model through joint optimization of latency and alignment.

\clearpage

\bibliography{refs}
\bibliographystyle{iclr2027_conference}

\clearpage

\appendix
\section*{Appendix}

\section{Benchmark Dataset Details}
\label{app:speed}

\looseness-1We use 798 samples from the SPEED Bench speculative decoding benchmark to evaluate all methods across 11 diverse domains: Coding (Code), Humanities (Hum.), Mathematics (Math), Multilingual (Multilin.), Question Answering (QA), Retrieval-Augmented Generation (RAG), Reasoning (Reason.), Roleplay (Role.), STEM, Summarization (Summ.), and Writing. The benchmark covers a broad range of generation tasks, enabling a comprehensive assessment of speculative decoding performance under different workloads. For our standard speculative decoding evaluation, we use a draft length of 4, following~\citet{chen2023accelerating}. Table~\ref{tab:speedbench_stats} summarizes the number of evaluation samples in each domain.

\begin{table}[h]
\centering
\small
\caption{Number of evaluation samples from SPEED Bench used in our experiments.}\vspace{3mm}
\label{tab:speedbench_stats}
\resizebox{\textwidth}{!}{
\begin{tabular}{lcccccccccccc}
\toprule
\makecell{\textbf{Code}}
& \makecell{\textbf{Hum.}}
& \makecell{\textbf{Math}}
& \makecell{\textbf{Multilin.}}
& \makecell{\textbf{QA}}
& \makecell{\textbf{RAG}}
& \makecell{\textbf{Reason.}}
& \makecell{\textbf{Role.}}
& \makecell{\textbf{STEM}}
& \makecell{\textbf{Summ.}}
& \makecell{\textbf{Writing}}
& \makecell{\textbf{Total}} \\
\midrule
80 & 80 & 80 & 80 & 80 & 71 & 51 & 48 & 80 & 78 & 70 & 798 \\
\bottomrule
\end{tabular}
}
\end{table}

\section{Training Dataset}
\label{app:training-dataset}

As discussed in Section~\ref{sec:results}, we use queries from open-source instruction-tuning datasets and generate responses using the target models. We randomly sample 250K queries, which cover approximately 95\% of the Llama tokenizer vocabulary. In total, these queries correspond to approximately 0.022B Llama tokens. Although the same set of queries is used for all target models, the number and length of the generated responses vary across models. To characterize the scale of the resulting training data, Table~\ref{tab:train-data} reports the total number of generated response tokens and the average response length for each target model.

\begin{table}[h]
\centering
\caption{Statistics of the generated training data. Both models are prompted with the same set of 250K instruction-tuning queries.}\vspace{3mm}
\label{tab:train-data}
\small
\begin{tabular}{lcc}
\toprule
\textbf{Target Model} & \textbf{Total Response Tokens (B)} & \textbf{Avg. Response Length} \\
\midrule

Llama-2-7B-Chat & 0.147 & 588 \\
Vicuna-13B-v1.5 & 0.144 & 577 \\
Qwen2.5-7B-Instruct & 0.142 & 567 \\
Qwen3-14B & 0.144 & 577 \\
Nemotron-32B & 0.147 & 588 \\
\bottomrule
\end{tabular}
\end{table}


\section{Implementation Details}
\label{app:imp-details}

We explain our implementation details briefly below.

\textbf{Baselines.} We use the \textit{Wide} model proposed by \citet{yan2025decoding} as our primary baseline. The \textit{Wide} architecture is obtained by pruning the target model, and we refer to this approach as \textit{Wide} throughout the paper. Specifically, we use the best-performing \textit{Wide} architecture reported by \citet{yan2025decoding}, which was optimized on Llama-2-7B-Chat. We then prune each target model according to this architecture and train it on the same dataset used for \texttt{T2D}. Following pruning, the resulting \textit{Wide} models are further fine-tuned using the same supervised fine-tuning dataset. For consistency, we also adopt the same 5-layer \textit{Wide} architecture proposed by \citet{yan2025decoding} across all target models. Similar to \texttt{T2D}, the \textit{Wide} models are optimized for two epochs and then continually fine-tuned for one additional epoch after pruning. We also compare our method with a simpler ShortGPT-based draft-model construction method with the same architecture as the \textit{Wide} model.

\textbf{Hyperparameters.} We summarize the hyperparameters used in \texttt{T2D} model development for Llama-2-7B-Chat model in Table \ref{tab:hyperparameters}. The other models use similar parameters, which we recommend tuning based on the model architecture.

\textbf{Hardware Configurations.} The experiments were conducted on a Linux server with 14x
NVIDIA DGX B200 GPUs with 192 GB VRAM/GPU.

\begin{table}[h]
    \centering
    \caption{Hyperparameters used for pruning and fine-tuning in \texttt{T2D} framework for Llama-2-7B-Chat.}\vspace{3mm}
    \label{tab:hyperparameters}
    \resizebox{0.7\linewidth}{!}{
    \begin{tabular}{lc}
        \toprule
        \textbf{Hyperparameter} & \textbf{Value} \\
        \midrule
        Target modules & Layers, attention heads, FFN channels \\
        Reference CE loss ($\mathcal{L}'_{CE}$) & $0.5$ \\
        Cross-entropy loss weight ($\mu_{CE}$) & $1.0$ \\
        Latency loss weight ($\mu_{lat}$) & $200$ \\
        Joint optimization epochs & $2$ \\
        Continual fine-tuning epochs & $1$ \\
        Global training batch size & $32$ \\
        Sequence length & $4096$ \\
        Fine-tuning learning rate & $1\times10^{-4}$ \\
        Layer gate learning rate & $1\times10^{-3}$ \\
        Head gate learning rate & $1\times10^{-3}$ \\
        FFN gate learning rate & $3\times10^{-3}$ \\
        Penalty learning rate ($\eta_1$) & $1\times10^{-3}$ \\
        Penalty learning rate ($\eta_2$) & $1\times10^{-3}$ \\
        \bottomrule
    \end{tabular}
    }
\end{table}

\section{\texttt{T2D}-Derived Draft Model Architectures}
\label{app:archi}
\label{app:model-architecture}

Table~\ref{tab:model-architecture} presents the final draft-model architectures derived by \texttt{T2D} for all \texttt{T2D} models.

\begin{table}[h]
\centering\vspace{-4mm}
\caption{Architectural specifications of the \texttt{T2D}-derived draft models for all \texttt{T2D} models.}\vspace{5pt}
\label{tab:model-architecture}
\resizebox{0.82\textwidth}{!}{
\begin{tabular}{lccccc}
\toprule
\textbf{\texttt{T2D} Draft Model} & \textbf{Params (B)} & \textbf{Layers} & \textbf{Hidden Size} & \textbf{Num Heads} & \textbf{FFN Size} \\
\midrule
Llama-2-7B-Chat & 0.86 & 3 & 4096 & 32 & 10752 \\

Vicuna-13B-v1.5 & 1.21 & 4 & 4480 & 35 & 11264 \\

Qwen2.5-7B-Instruct & 1.54 & 2 & 3584 & 28 & 18432 \\

Qwen3-14B & 2.20 & 2 & 5120 & 40 & 16896 \\

Nemotron-32 & 2.51 & 2 & 5120 & 40 & 27136 \\

\bottomrule
\end{tabular}
}
\end{table}

\section{Ratio of Weights, CE Loss, and Latency over the Training Period for other models}
\label{app:trend-others}

\begin{figure*}[h]
    \centering
    \includegraphics[width=0.5\linewidth]{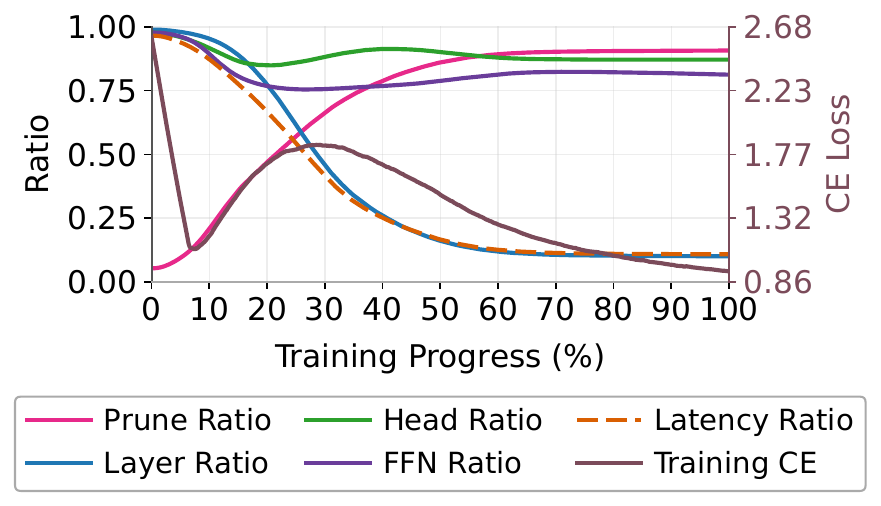}
    \caption{Pruning ratio of the target model (Vicuna-13B-v1.5) over the training period, along with the kept percentage (ratio) of the Layers, Heads, and FFN. We also plot latency and CE loss.}
    \label{fig:vicuna-trend}
\end{figure*}

\begin{figure*}[h]
    \centering
    \includegraphics[width=0.5\linewidth]{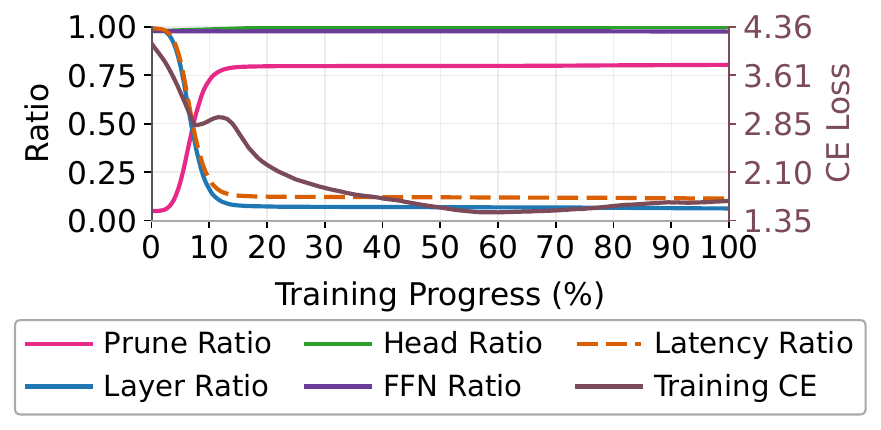}
    \caption{Pruning ratio of the target model (Qwen2.5-7B-Instruct) over the training period, along with the kept percentage (ratio) of the Layers, Heads, and FFN. We also plot latency and CE loss.}
    \label{fig:palm2-trend}
\end{figure*}

\begin{figure*}[h]
    \centering
    \includegraphics[width=0.5\linewidth]{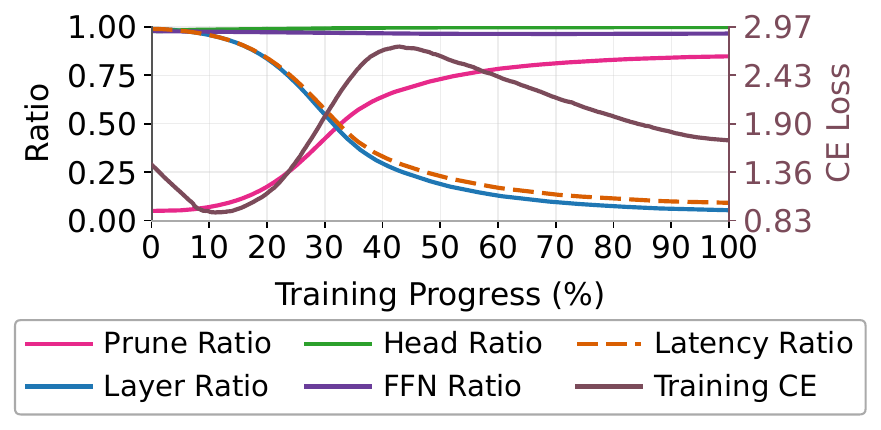}
    \caption{Pruning ratio of the target model (Qwen3-14B) over the training period, along with the kept percentage (ratio) of the Layers, Heads, and FFN. We also plot latency and CE loss.}
    \label{fig:palm3-trend}
\end{figure*}

\begin{figure*}[h]
    \centering
    \includegraphics[width=0.5\linewidth]{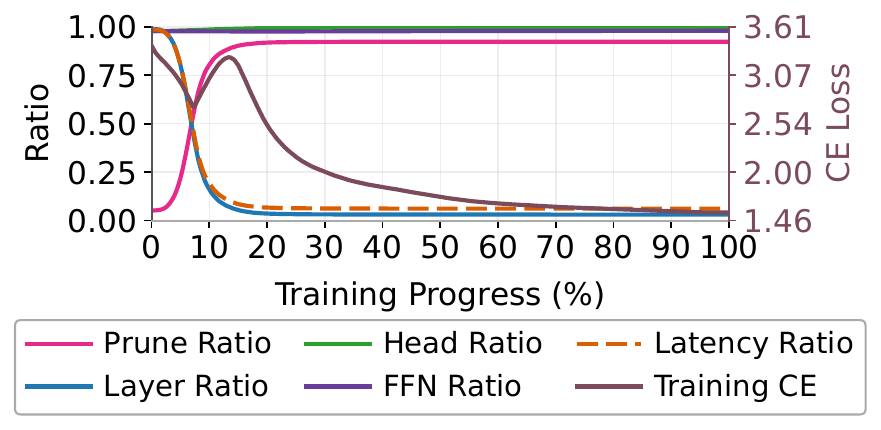}
    \caption{Pruning ratio of the target model (Nemotron-32B) over the training period, along with the kept percentage (ratio) of the Layers, Heads, and FFN. We also plot latency and CE loss.}
    \label{fig:nemotron-trend}
\end{figure*}

We provide the ratios of different components of other models during the \texttt joint optimization in Figures~\ref{fig:vicuna-trend},~\ref{fig:palm2-trend},~\ref{fig:palm3-trend} and~\ref{fig:nemotron-trend}. Figure~\ref{fig:vicuna-trend} shows that for Vicuna-13B-v1.5, the estimated latency decreases over the training period, along with the ratio of the pruned model size to the target model size. However, when examining individual components, we observe that the number of attention heads in Vicuna-13B-v1.5 initially decreases due to pruning and later increases slightly as some heads are retained. For the FFN, the ratio initially decreases and eventually flattens. These trends show that \texttt{T2D} can automatically adjust the pruning ratios of different components during optimization. For Qwen2.5-7B-Instruct, Qwen3-14B, and Nemotron-32B, the ratios follow similar trends over the training periods.

\section{Per Domain Speedup Comparison with Sheared Llama}
\label{app:per-domain-sheared}

Table \ref{tab:domain-speed-shear} shows the speedup of Sheared-Llama-1.3B and the \texttt{T2D} draft model across domains. \texttt{T2D} achieves higher speedup in every domain because of the latency-aware optimization in \texttt{T2D}.

\begin{table*}[h]
\centering
\caption{Speculative decoding speedup ($\times$) of Sheared-Llama-1.3B and \texttt{T2D}. 
The highest speedup for each domain is shown in \textbf{bold}.}\vspace{3mm}
\label{tab:domain-speed-shear}
\fontsize{8}{9}\selectfont
\setlength{\tabcolsep}{3.5pt}
\resizebox{\textwidth}{!}{
\begin{tabular}{llccccccccccccc}
\toprule
\textbf{Draft Model} & \textbf{Size (B)}
& \makecell{\textbf{Code}}
& \makecell{\textbf{Hum.}}
& \makecell{\textbf{Math}}
& \makecell{\textbf{Multilin.}}
& \makecell{\textbf{QA}}
& \makecell{\textbf{RAG}}
& \makecell{\textbf{Reason.}}
& \makecell{\textbf{Role.}}
& \makecell{\textbf{Stem}}
& \makecell{\textbf{Summ.}}
& \makecell{\textbf{Writing}}
& \makecell{\textbf{Overall}} \\
\midrule

Sheared-Llama-1.3B-ft & 1.27
& 1.32 & 1.11 & 1.29 & 1.16 & 0.90 & 1.36 & 1.40
& 1.12 & 1.20 & 0.73 & 1.27 & 1.18 \\
\texttt{T2D} & 0.86
& 1.98 & 1.59 & 1.92 & 2.03 & 1.46 & 1.79 & 1.90
& 1.35 & 1.72 & 1.60 & 1.62 & 1.74 \\

\bottomrule
\end{tabular}
}
\end{table*}




\section{Development of the MLP Latency Predictor}
\label{app:mlp-lat-pred}

To obtain a differentiable estimate of decoding latency, we train a lightweight MLP latency predictor offline before the joint pruning-and-distillation optimization. The predictor takes the estimated architectural dimensions $(L_D, H_D, I_D)$, corresponding to the retained layers, attention heads, and FFN size, and predicts the log latency of a single cached decoding step.

\paragraph{Training Data.}
We construct a dataset of candidate architectures by varying the number of layers, attention heads, and FFN dimensions. For each architecture, we instantiate an uninitialized model of the corresponding shape and directly measure single-token decoding latency on the target hardware. The median latency across repeated measurements is used as the regression target.

\paragraph{Model and Training.}
The latency predictor is a three-layer MLP with two hidden layers of size 256 and SiLU activations. Inputs are standardized using training-set statistics, and the model is trained to predict log latency using the Huber loss~\citep{gokcesu2021generalized}. To encourage smooth interpolation between measured architectures, we apply small random perturbations to the input dimensions during training. After training, the predictor is frozen and used throughout the joint optimization process to provide differentiable latency estimates.

\section{Qualitative Analysis of Speculative Decoding Performance}
\label{app:sd-performance}

\textbf{Text generation quality of \texttt{T2D}.}As we have already discussed, \texttt{T2D} prunes the target model by up to $92\%$. Following this extensive pruning and alignment, we investigate whether the resulting draft model retains coherent text-generation capability. Table~\ref{tab:t2d-responses} provides representative responses generated by the \texttt{T2D} Llama-2 draft model for a range of random general-purpose queries. Most of the time, the model is successful in staying relevant to the query and topic, correctly following instructions and producing on-topic, coherent answers.

\begin{table}[t]
\centering
\caption{Example response generated by the \texttt{T2D} Llama-2 draft model, with observations on response quality.}
\label{tab:t2d-responses}
\small
\begin{tabular}{p{0.18\linewidth} p{0.46\linewidth} p{0.26\linewidth}}
\toprule
\textbf{Prompt} & \textbf{\texttt{T2D} Draft Model Response (excerpt)} & \textbf{Observation} \\
\midrule
How are you? &
I'm just an AI, I don't have personal experiences or emotions, so I don't have a physical body or a specific situation. However, I can provide information on a wide range of topics, including but not limited to: 1. Travel and exploration: I can provide information on different cultures, languages, and landscapes, and I can also provide information on different cultures and landscapes. 2. Culture and traditions: I can provide information on different cultures and traditions [...] &
Relevant and appropriate framing (correctly disclaims lacking emotions) repetition. \\
\addlinespace
What is the capital of France? &
The capital of France is Paris. &
Fully correct and relevant; concise factual answer. \\
\addlinespace
Can you help me in my study? &
Of course! I'd be happy to help you in your study. Here are some steps you can take to help me improve my study skills: 1. Identify the type of study: Determine which type of study you want to focus on, such as studies that you want to study for a specific purpose or subject. 2. Research the topic: Identify the topic or subject you want to focus on, such as the impact of social media on mental health [...] &
Relevant and largely coherent, giving actionable study advice. \\
\addlinespace
What is a computer? &
A computer is a type of computer that is designed to perform specific tasks, such as image processing, data analysis, and machine learning. It is typically used in a variety of applications, including: 1. Image processing: A computer can process images, such as images of objects, scenes, or time-series data, and can be used to perform tasks such as image compression, feature extraction, and image processing [...] &
Topically relevant (stays within computing/AI applications) but the core definition is circular. \\
\addlinespace
What is Artificial Intelligence? &
Artificial Intelligence (AI) is a rapidly growing field that has the potential to revolutionize various industries, including healthcare, finance, transportation, and education. Here are some of the key aspects of AI: 1. Machine Learning: AI systems can be used to analyze large datasets, identify patterns, and make predictions, and optimize them for better decision-making. 2. Natural Language Processing (NLP): AI systems can be used to analyze large amounts of text data, such as customer reviews, marketing campaigns, and customer feedback, to understand their context and make more informed decisions [...] &
Correct and directly relevant; accurately defines AI and enumerates coherent, describes the subfields (ML, NLP), demonstrating strong topical grounding. \\

\bottomrule
\end{tabular}
\end{table}

\textbf{Comparison with baselines on Standard Speculative Decoding.} As discussed in the previous section, \texttt{T2D} generates coherent and meaningful text, which is important for a draft model because its predictions are used to propose tokens during speculative decoding. We now compare the token sequences generated by \texttt{T2D}, ShortGPT, and \textit{Wide}, with all three evaluated against the same target model. Figure~\ref{fig:qualitative-traces} shows decoding traces from three domains: RAG-style extraction, reasoning, and open-ended writing. Accepted tokens are shown in \acc{green}, rejected tokens in \rej{red}, and the target model's corrections in \cor{blue}.

Across the examples, the three draft models often make similar mistakes early in the generation. For example, they may propose \texttt{The} instead of \texttt{Based} at the beginning of a response, or \texttt{text} instead of \texttt{data}/\texttt{question}. These cases suggest that some errors are shared across the pruned draft models rather than being specific to a particular construction method. At the same time, \texttt{T2D} gets several finer details right where the baselines do not. In Example~1, for instance, \texttt{T2D} correctly predicts the capitalized \texttt{D} in \texttt{D party}, while both ShortGPT and \textit{Wide} predict the lowercase \texttt{d}. In Example~3, \texttt{T2D} behaves similarly to ShortGPT and maintains a long accepted span, while \textit{Wide} introduces several additional rejections even though it eventually produces similar text. Overall, these traces show that \texttt{T2D} can maintain fluent generation and capture details such as capitalization and entity-level tokens, while still exhibiting some of the common errors shared by other pruned draft models.

\begin{figure*}[t]
\small
\raggedright

\begin{tcolorbox}[colback=white,colframe=boxBorder,boxrule=0.6pt,arc=1.5mm,left=6pt,right=6pt,top=5pt,bottom=5pt]
\textbf{Example 1: Context Tabular Extraction (RAG)} \\
\textit{Target:} ``Based on the provided data, the names of the representatives from the D party are:''
\vspace{3pt}\hrule\vspace{5pt}

\noindent\textbf{T2D (Ours)} \\
\rej{The}\arrow\cor{Based} \acc{on the} \rej{information}\arrow\cor{provided} \rej{text}\arrow\cor{data,} \acc{the names of the representatives from the D party are:}
\vspace{4pt}

\noindent\textbf{ShortGPT} \\
\rej{The}\arrow\cor{Based} \acc{on the} \rej{given}\arrow\cor{provided} \rej{text}\arrow\cor{data,} \acc{the} \rej{represent}\arrow\cor{names} \acc{of the representatives from the} \rej{d}\arrow\cor{D} \acc{party are:}
\vspace{4pt}

\noindent\textbf{\textit{Wide}} \\
\rej{The}\arrow\cor{Based} \acc{on the provided} \rej{text}\arrow\cor{data,} \acc{the} \rej{represent}\arrow\cor{names} \acc{of the representatives from the} \rej{d}\arrow\cor{D} \acc{party are:}
\end{tcolorbox}
\vspace{-2pt}

\begin{tcolorbox}[colback=white,colframe=boxBorder,boxrule=0.6pt,arc=1.5mm,left=6pt,right=6pt,top=5pt,bottom=5pt]
\textbf{Example 2: Multi-Step Reasoning} \\
\textit{Target:} ``Based on the provided question, the answer is (E) No, the jury's award is discretionary...''
\vspace{3pt}\hrule\vspace{5pt}

\noindent\textbf{T2D (Ours)} \\
\rej{The}\arrow\cor{Based} \acc{on the} \rej{information}\arrow\cor{provided} \rej{text}\arrow\cor{question,} \acc{the answer is (} \rej{X}\arrow\cor{E} \acc{) No, the} \rej{passage}\arrow\cor{jury's} \acc{award is} \rej{not}\arrow\cor{discretionary}
\vspace{4pt}

\noindent\textbf{ShortGPT} \\
\rej{The}\arrow\cor{Based} \acc{on the} \rej{information}\arrow\cor{provided} \rej{text}\arrow\cor{question,} \acc{the answer is (} \rej{B}\arrow\cor{E} \acc{) No,} \rej{because}\arrow\cor{the} \rej{app}\arrow\cor{jury's} \acc{award is} \rej{not}\arrow\cor{discretionary}
\vspace{4pt}

\noindent\textbf{\textit{Wide}} \\
\rej{The}\arrow\cor{Based} \acc{on the} \rej{given}\arrow\cor{provided} \rej{text}\arrow\cor{question,} \acc{the answer} \rej{to}\arrow\cor{is} \acc{(} \rej{X}\arrow\cor{E} \acc{)} \rej{Yes}\arrow\cor{No,} \acc{the} \rej{evidence}\arrow\cor{jury's} \acc{award is} \rej{not}\arrow\cor{discretionary}
\end{tcolorbox}
\vspace{-2pt}

\begin{tcolorbox}[colback=white,colframe=boxBorder,boxrule=0.6pt,arc=1.5mm,left=6pt,right=6pt,top=5pt,bottom=5pt]
\textbf{Example 3: Pedagogical Writing (Open-Ended)} \\
\textit{Target:} ``Sure, I'd be happy to help you write a lesson plan for a kindergarten middle class activity themed "I'm a Little Scientist"...''
\vspace{3pt}\hrule\vspace{5pt}

\noindent\textbf{T2D (Ours)} \\
\rej{This}\arrow\cor{Sure} \acc{,} \rej{here}\arrow\cor{I} \acc{'d be happy to help you write a lesson plan for a kindergarten middle class activity themed "I'm a Little Scientist"}
\vspace{4pt}

\noindent\textbf{ShortGPT} \\
\rej{This}\arrow\cor{Sure} \acc{,} \rej{here}\arrow\cor{I} \acc{'d be happy to help you write a lesson plan for a kindergarten middle class activity themed "I'm a Little Scientist"}
\vspace{4pt}

\noindent\textbf{\textit{Wide}} \\
\acc{Sure,} \rej{here}\arrow\cor{I} \rej{can}\arrow\cor{'} \acc{d be happy to help you} \rej{with}\arrow\cor{write} \acc{a lesson plan for a kindergarten middle class activity} \rej{!}\arrow\cor{themed} \rej{by}\arrow\cor{"} \acc{I'm a Little Scientist"}
\end{tcolorbox}

\vspace{-4pt}
\caption{Examples of speculative decoding outputs across different domains.}
\label{fig:qualitative-traces}
\end{figure*}

\end{document}